\documentclass[11pt]{article}

\usepackage[final]{acl}

\usepackage{times}
\usepackage{latexsym}

\usepackage[T1]{fontenc}

\usepackage[utf8]{inputenc}

\usepackage{microtype}

\usepackage{inconsolata}

\usepackage{graphicx}
\usepackage{amssymb}  

\usepackage{algorithm}
\usepackage{algorithmicx}
\usepackage{algpseudocode}
\usepackage{tabularray}
\usepackage{caption}
\usepackage{subcaption}
\usepackage{amsmath}
\usepackage{multirow}

\title{BnTTS: Few-Shot Speaker Adaptation in Low-Resource Setting}

\author{
\textbf{Mohammad Jahid Ibna Basher}\textsuperscript{1},
\textbf{Md Kowsher}\textsuperscript{2},
\textbf{Md Saiful Islam}\textsuperscript{1},
\textbf{Rabindra Nath Nandi}\textsuperscript{1}, \\
\textbf{Nusrat Jahan Prottasha}\textsuperscript{2},
\textbf{Mehadi Hasan Menon}\textsuperscript{1},
\textbf{Tareq Al Muntasir}\textsuperscript{1}, \\
\textbf{Shammur Absar Chowdhury}\textsuperscript{3},
\textbf{Firoj Alam}\textsuperscript{3},
\textbf{Niloofar Yousefi}\textsuperscript{2}, 
\textbf{Ozlem Ozmen Garibay}\textsuperscript{2} \\
\textsuperscript{1}Hishab Singapore Pte. Ltd, Singapore, 
\textsuperscript{2}University of Central Florida, USA \\
\textsuperscript{3}Qatar Computing Research Institute, Qatar
}

\begin{document}
\maketitle
\begin{abstract}

This paper introduces BnTTS (\textbf{B}a\textbf{n}gla \textbf{T}ext-\textbf{T}o-\textbf{S}peech), the first framework for Bangla speaker adaptation-based TTS, designed to bridge the gap in Bangla speech synthesis using minimal training data. Building upon the XTTS architecture, our approach integrates Bangla into a multilingual TTS pipeline, with modifications to account for the phonetic and linguistic characteristics of the language. We pretrain BnTTS on 3.85k hours of Bangla speech dataset with corresponding text labels and evaluate performance in both zero-shot and few-shot settings on our proposed test dataset. Empirical evaluations in few-shot settings show that BnTTS significantly improves the naturalness, intelligibility, and speaker fidelity of synthesized Bangla speech. Compared to state-of-the-art Bangla TTS systems, BnTTS exhibits superior performance in Subjective Mean Opinion Score (SMOS), Naturalness, and Clarity metrics.

\end{abstract}

\section{Introduction}
\label{sec:introduction}

Speaker adaptation in Text-to-Speech (TTS) technology has seen substantial advancements in recent years, particularly with speaker-adaptive models enhancing the naturalness and intelligibility of synthesized speech \cite{eren2023deep}. Notably, recent innovations have emphasized zero-shot and one-shot adaptation approaches \cite{kodirov2015unsupervised}. Zero-shot TTS models eliminate the need for speaker-specific training by generating speech from unseen speakers using reference audio samples \cite{min2021meta}. Despite this progress, zero-shot models often require large datasets and face challenges with out-of-distribution (OOD) voices, as they struggle to adapt effectively to novel speaker traits \cite{le2023voicebox, ju2024naturalspeech3}. Alternatively, one-shot adaptation fine-tunes pre-trained models using a single data instance, offering improved adaptability with reduced data and computational demands \cite{yan2021adaspeech2, wang2023neural}; however, the pretraining stage still necessitates substantial datasets \cite{zhang2021transfer}.

Recent works such as YourTTS \cite{bai2022yourtts} and VALL-E X \cite{xu2022vall} have made strides in cross-lingual zero-shot TTS, with YourTTS exploring English, French, and Portuguese, and VALL-E X incorporating language identification to extend support for a broader range of languages \cite{xu2022vall}. These advancements highlight the potential for multilingual TTS systems to achieve cross-lingual speech synthesis. Furthermore, the XTTS model \cite{casanova2024xtts} represents a significant leap by expanding zero-shot TTS capabilities across 16 languages. Based on the Tortoise model \cite{casanova2024xtts}, XTTS enhances voice cloning accuracy and naturalness but remains focused on high- and medium-resource languages, leaving low-resource languages such as Bangla underserved \cite{zhang2022universal, xu2023cross}. 

The scarcity of extensive datasets has hindered the adaptation of state-of-the-art (SOTA) TTS models for low-resource languages. Models like YourTTS \cite{bai2022yourtts}, VALL-E X \cite{baevski2022vall}, and Voicebox \cite{baevski2022voicebox} have demonstrated success in multilingual settings, yet their primary focus remains on languages with rich resources like English, Spanish, French, and Chinese. While a few Bangla TTS systems exist \cite{gutkin2016tts}, they often produce robotic tones \cite{hossain2018development} or are limited to a small set of static speakers \cite{gong2024initial}, lacking instant speaker adaptation capabilities and typically not being open-source.

To address these challenges, we propose the first framework for few-shot speaker adaptation in Bangla TTS. Our approach integrates Bangla into the XTTS training pipeline, with architectural modifications to accommodate Bangla’s unique phonetic and linguistic features. Our model is optimized for effective few-shot voice cloning, addressing the needs of low-resource language settings. \textbf{Our contributions} are summarized as follows: \textbf{(i)} we present the \textit{first} speaker-adapted Bangla TTS system; \textbf{(ii)} we integrate Bangla into a multilingual XTTS pipeline, optimizing the framework to accommodate the unique challenges of low-resource languages; \textbf{(iii)} we make the developed BnTTSTextEval evaluation dataset public.




\section{BnTTS}
\begin{figure}[ht!]
    \raggedleft
    \includegraphics[width=0.90\linewidth]{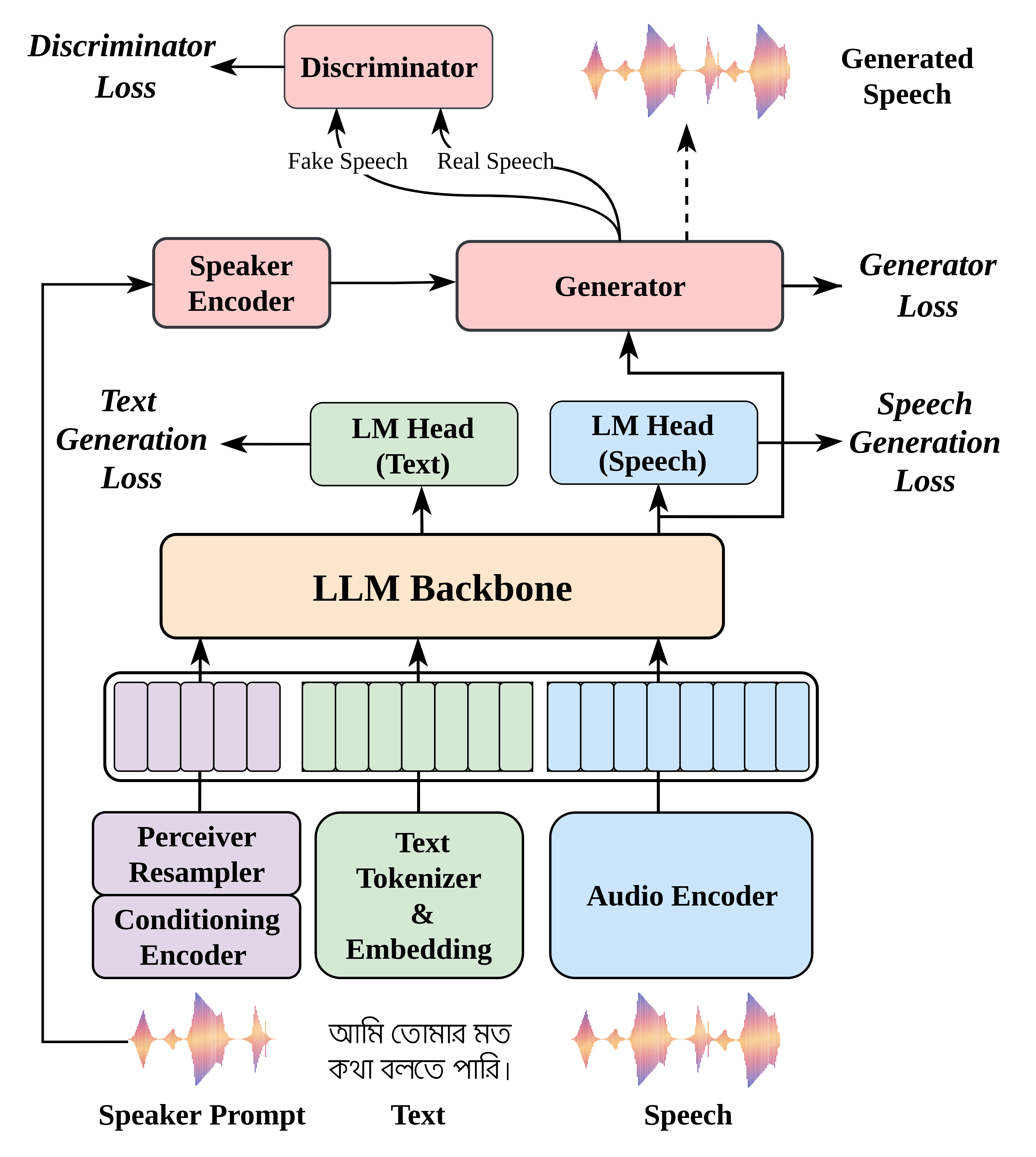} 
    \caption{Overview of BnTTS Model.} 
    \label{fig:xtts_train_diagram}
\end{figure}

\textbf{Preliminaries:} Given a text sequence with \( N \) tokens, \( \mathbf{T} = \{t_1, t_2, \ldots, t_N\} \), and a speaker's mel-spectrogram \( \mathbf{S} = \{s_1, s_2, \ldots, s_L\} \), the objective is to generate speech \( \hat{\mathbf{Y}} \) that matches the speaker's characteristics. The ground truth mel-spectrogram frames for the target speech are denoted as \( \mathbf{Y} = \{y_1, y_2, \ldots, y_M\} \). The synthesis process can be described as:
\[
\hat{\mathbf{Y}} = \mathcal{F}(\mathbf{S}, \mathbf{T})
\]
where \( \mathcal{F} \) produces speech conditioned on both the text and the speaker's spectrogram.

\noindent \textbf{Audio Encoder:} A Vector Quantized-Variational AutoEncoder (VQ-VAE) \cite{tortoise} encodes mel-spectrogram frames \( \mathbf{Y} \) into discrete tokens \( M \in \mathcal{C} \), where $\mathcal{C}$ is vocab or codebook. An embedding layer then transforms these tokens into a \( d \)-dimensional vector: \( \mathbf{Y_e} \in \mathbb{R}^{M \times d} \).

\noindent \textbf{Conditioning Encoder \& Perceiver Resampler:} The Conditioning Encoder \cite{casanova2024xtts} consists of \( l \) layers of \( k \)-head Scaled Dot-Product Attention, followed by a Perceiver Resampler. The speaker spectrogram \( \mathbf{S} \) is transformed into an intermediate representation \( \mathbf{S_z} \in \mathbb{R}^{L \times d} \), where each attention layer applies a scaled dot-product attention mechanism. The Perceiver Resampler generates a fixed output dimensionality \( \mathbf{R} \in \mathbb{R}^{P \times d} \) from a variable input length \( L \).

\noindent \textbf{Text Encoder:} The text tokens \( \mathbf{T} = \{t_1, t_2, \ldots, t_N\} \) are projected into a continuous embedding space, yielding \( \mathbf{T_e} \in \mathbb{R}^{N \times d} \).

\noindent \textbf{Large Language Model (LLM):} The transformer-based LLM \cite{radford2019language} utilizes the decoder portion. Speaker embeddings \( \mathbf{S_p} \), text embeddings \( \mathbf{T_e} \), and ground truth spectrogram embeddings \( \mathbf{Y_e} \) are concatenated to form the input:
\[
\mathbf{X} = \mathbf{S_p} \oplus \mathbf{T_e} \oplus \mathbf{Y_e} \in \mathbb{R}^{(N + P + M) \times d}
\]
The LLM processes \( \mathbf{X} \), producing output \( \mathbf{H} \) with hidden states for the text, speaker, and spectrogram embeddings. During inference, only text and speaker embeddings are concatenated, generating spectrogram embeddings \( \{h_1^Y, h_2^Y, \ldots, h_P^Y\} \) as the output.

\noindent \textbf{HiFi-GAN Decoder:} The HiFi-GAN Decoder \cite{kong2020hifi} converts the LLM's output into realistic speech, preserving the speaker's characteristics. Specifically, it takes the LLM's speech head output \( \mathbf{H}_\text{Y} = \{h_1^Y, h_2^Y, \ldots, h_P^Y\} \). The speaker embedding \( \mathbf{S} \) is resized to match \( \mathbf{H}_\text{Y} \), resulting in \( \mathbf{S}' \in \mathbb{R}^{P \times d} \). The final audio waveform \( \mathbf{W} \) is then generated by:
\[
\mathbf{W} = g_\text{HiFi}(\mathbf{H}_\text{Y} + \mathbf{S}')
\]
Thus, the HiFi-GAN decoder produces speech that reflects the input text while maintaining the speaker's unique qualities.

\section{Experiments}

\noindent \textbf{BnTTS model:} BnTTS employs the pretrained XTTS checkpoint \cite{casanova2024xtts} as its base model, chosen for resource efficiency. The Conditioning Encoder has six attention blocks with 32 heads, capturing contextual information. The Perceiver Resampler reduces the sequence to a fixed length of 32. The model maintains GPT-2’s dimensionality, with a hidden size of 1024 and an intermediate layer size of 3072, handling sequences of up to 400 tokens. (Details in Appendix \ref{app:training_objective}).

\noindent \textbf{Dataset:} We continuously pre-trained the BnTTS model (initialized from the XTTS checkpoint) on 3.85k hours of Bengali speech data, sourced from open-source datasets, pseudo-labeled data, and synthetic datasets. The pseudo-labeled data were collected using an in-house automated TTS Data Acquisition Framework, which segments speech into 0.5 to 11-second chunks with time-aligned transcripts. These segments were further refined using neural speech models and custom algorithms to enhance quality and accuracy. For speaker adaptation, we incorporated 4.22 hours of high-quality studio recordings from four speakers, referred to as In-House HQ Data.

For evaluation, we propose two datasets: (1) BnStudioEval, derived from our In-House HQ Data, to assess high-fidelity speech generation and speaker adaptation, and (2) BnTTSTextEval, a text-only dataset consisting of three subsets: BengaliStimuli53 (assessing phonetic diversity), BengaliNamedEntity1000 (evaluating named entity pronunciation), and ShortText200 (measuring conversational fluency in short sentences, filler words, and common phrases used in everyday dialogue). Further details are provided in Appendices \ref{sec:data_collection}, \ref{app:human_reviewed_data}, and \ref{app:dataset}.

\noindent \textbf{Training Setup:} We initialized the BnTTS model from the XTTS checkpoint and do continual pre-training using the AdamW optimizer with betas of 0.9 and 0.96, weight decay of 0.01, and an initial learning rate of 2e-05. The batch size was 12, with gradient accumulation over 24 steps per GPU, and the learning rate decay(0.66) was applied using MultiStepLR. All experiments are run on a single NVIDIA A100 GPU with 80GB of VRAM. The pretraining process consists of two stages:

\textbf{a) Partial Audio Prompting:} In this stage, a random segment of the ground truth audio is used as the speaker prompt. Training in this phase lasted for 5 epochs.

\textbf{b) Complete Audio Prompting:} Here, the full duration of audio is used as the speaker prompt. This stage continues from the checkpoint and optimizer state of the first phase and lasts for 1 epoch.

Additionally, the HiFi-GAN vocoder was fine-tuned separately using GPT-2 embeddings derived from the model in stage b. The vocoder was fine-tuned for three days to ensure optimal performance. The audio encoder and speaker encoder remain frozen across all experiments.

\noindent \textbf{Few-shot Speaker Adaptation:} For few-shot speaker adaptation, we fine-tuned the BnTTS model using our In-House HQ dataset, which comprises studio recordings from four speakers. We randomly selected 20 minutes of audio for each speaker and fine-tuned the model in a multi-speaker setting for 10 epochs. This fine-tuning approach is more meaningful with the XTTS-like architecture pretrained on large-scale datasets. The evaluation results are presented in Section \ref{sec:results}.



\noindent \textbf{Evaluation Metric: } We evaluate the BnTTS system using six criteria. The Subjective Mean Opinion Score (SMOS) including Naturalness and Clarity evaluates perceived audio quality from \citet{streijl2016mean}, while the ASR-based Character Error Rate (CER) \cite{nandi-etal-2023-pseudo} measures transcription accuracy, SpeechBERTScore assesses similarity to reference speech, and Speaker Encoder Cosine Similarity (SECS) evaluates speaker identity fidelity \cite{saeki2024spbertscore, casanova2021sc, thienpondt2024ecapa2}. See Appendix \ref{app:eval_metrics} for details.

\section{Results}
\label{sec:results}
We evaluated the pretrained BnTTS (BnTTS-0) and speaker-adapted BnTTS (BnTTS-n) alongside IndicTTS \cite{indictts2022} and two commercial systems: Google Cloud TTS (GTTS) and Azure TTS (AzureTTS). The evaluation was conducted on both the BnStudioEval and BnTTSTextEval datasets. For a time-efficient subjective evaluation, we randomly selected 200 sentences from the BengaliNamedEntity1000 subset, which originally contains 1000 samples, maintaining a comprehensive assessment while reducing evaluation overhead.

\noindent \textbf{ Reference-aware Evaluation:}
Table \ref{tab:eval_on_studio_eval} shows the performance of various TTS systems on the BnStudioEval dataset. GTTS outperforms other methods in the CER metric, even surpassing the Ground Truth (GT) in transcription accuracy. As for the subjective measures, the proposed BnTTS-n closely follows the GT, with competitive scores in SMOS (4.624 {\em vs} 4.809), Naturalness (4.600 {\em vs} 4.798), and Clarity (4.869 {\em vs} 4.913). Meanwhile, BnTTS-0 achieves SMOS, Naturalness, and Clarity scores of 4.456, 4.447, and 4.577, respectively. IndicTTS, AzureTTS, and GTTS perform poorly in the subjective metrics.

In speaker similarity evaluation, GT attains a perfect SECS (reference) score and high SECS (prompt) scores. BnTTS-n outperforms BnTTS-0 in both SECS (reference) (0.548 {\em vs} 0.529) and SECS (prompt) (0.586 {\em vs} 0.576). Additionally, BnTTS-n achieves a SpeechBERTScore of 0.791, slightly higher than BnTTS-0 at 0.789, while GT retains a perfect score of 1.0. IndicTTS, GTTS, and AzureTTS do not support speaker adaptation, so SECS and SpeechBERTScore were not evaluated for these systems.

\noindent \textbf{ Reference-independent Evaluation:} Table \ref{tab:eval_on_BnTTSTextEval} presents the comparative performance of various TTS systems evaluated on the BnTTSTextEval dataset. The AzureTTS and GTTS consistently achieve lower CER scores, with BnTTS-n and BnTTS-0 following closely in third and fourth place, respectively, and IndicTTS trailing behind.
BnTTS-n performs strongly in subjective evaluations, excelling in SMOS, Naturalness, and Clarity scores across the BengaliStimuli53, BengaliNamedEntity1000, and ShortText200 subsets. Overall, BnTTS-n achieves the highest scores in SMOS (4.601), Naturalness (4.578), and Clarity (4.832). Meanwhile, AzureTTS performs competitively, surpassing other commercial and open-source models and achieving scores comparable to BnTTS-0.

\renewcommand{\arraystretch}{1.1} 
\begin{table}[ht!]
\centering
\footnotesize
\setlength{\tabcolsep}{2.9pt}
\resizebox{0.47\textwidth}{!}{%
\begin{tabular}{l|cccccc}
\hline
\textbf{Method} & \textbf{GT} & \textbf{IndicTTS} & \textbf{GTTS} & \textbf{AzureTTS}  &\textbf{BnTTS-0}& \textbf{BnTTS-n}\\ 
\hline
CER & \textit{0.030}& 0.058& \textbf{0.020}& 0.021&0.052& 0.034\\ 
SMOS & \textit{4.809}& 3.475& 4.017& 4.154&4.456& \textbf{4.624}\\ 
Naturalness & \textit{4.798}& 3.406& 3.949& 4.100&4.447& \textbf{4.600}\\ 
Clarity & \textit{4.913}& 4.160& 4.700& 4.686&4.577& \textbf{4.869}\\ 
SECS (Ref.) & \textit{1.0} & - & - & -  &0.529& \textbf{0.548}\\ 
SECS (Prompt) & \textit{0.641}& - & - & -  &0.576& \textbf{0.586}\\ 
SpeechBERT-\\ Score& \textit{1.0} & - & - & -  &0.789& \textbf{0.791} \\ 
\hline
\end{tabular}}
\caption{Comparative average performance for reference-aware BnStudioEval dataset. SECS and SpeechBERTScore are not reported for  IndicTTS, GTTS, and AzureTTS as these systems do not support speaker adaption.}
\label{tab:eval_on_studio_eval}
\vspace{-0.2cm}
\end{table}
\renewcommand{\arraystretch}{1.1} 
\begin{table}[ht!]
\centering
\footnotesize
\setlength{\tabcolsep}{3.5pt}
\resizebox{0.47\textwidth}{!}{%
\begin{tabular}{c|c|c c c c}
\hline
\textbf{Dataset} & \textbf{Method} & \textbf{CER} & \textbf{SMOS} & \textbf{Naturalness} & \textbf{Clarity} \\ 
\hline
Bengali- & IndicTTS & 0.110& 3.445& 3.403& 3.857\\ 
Stimuli-53& GTTS     & 0.063& 4.006& 3.937& 4.688\\ 
          & AzureTTS& \textbf{0.060}& 4.108& 4.064& 4.542\\
 & BnTTS-0& 0.092& 4.622& 4.613&4.719\\ 
          & BnTTS-n& 0.086& \textbf{4.654}& \textbf{4.634}& \textbf{4.854}\\ 
\hline
Bengali- & IndicTTS & 0.049& 3.527& 3.462& 4.179\\ 
Named-& GTTS & 0.037& 4.037& 3.969& 4.712\\ 
                 Entity-& AzureTTS& \textbf{0.032}& 4.182& 4.135& 4.654\\
 1000& BnTTS-0& 0.043& 4.585& 4.613&4.698\\ 
                 (200)& BnTTS-n& 0.040& \textbf{4.635}& \textbf{4.614}& \textbf{4.841}\\ 
\hline
Short- & IndicTTS & 0.204& 3.233& 3.325& 3.893\\ 
Text-200& GTTS   & \textbf{0.043}& 4.058& 3.993& 4.705\\ 
         & AzureTTS& 0.050& 4.294& 4.256& 4.675\\
 & BnTTS-0& 0.116& 4.297& 4.271&4.556\\ 
         & BnTTS-n& 0.092& \textbf{4.554}& \textbf{4.528}& \textbf{4.816}\\ 
\hline
Overall & IndicTTS & 0.125& 3.388& 3.325& 4.017\\ 
        & GTTS     & 0.049 & 4.042& 3.976& 4.706\\ 
        & AzureTTS& \textbf{0.045}& 4.223& 4.180& 4.650\\
 & BnTTS-0& 0.081& 4.463& 4.445&4.639\\ 
        & BnTTS-n& 0.069& \textbf{4.601}& \textbf{4.578}& \textbf{4.832}\\ 
\hline
\end{tabular}}
\caption{Comparative average performance analysis on the reference-independent BnTTSTextEval dataset.}
\label{tab:eval_on_BnTTSTextEval}
\vspace{-0.2cm}
\end{table}
\begin{table}[ht!]
\centering
\resizebox{0.4\textwidth}{!}{%
\begin{tabular}{c|cccc} 
\hline
\textbf{Exp.} & \textbf{T and TopK} & \begin{tabular}[c]{@{}c@{}}\textbf{Short}\\\textbf{Prompt}\end{tabular} & \begin{tabular}[c]{@{}c@{}}\textbf{Duration}\\\textbf{Equality}\end{tabular} & \textbf{CER}  \\ 
\hline
1                & T=0.85, TopK=50     & N                                                                       & 0.699                                                                        & 0.081         \\ 

2                & T=0.85, TopK=50     & Y                                                                       & 0.820                                                                        & 0.029         \\ 

3                & T=1.0, TopK=2       & N                                                                       & 0.701                                                                        & 0.023         \\ 

4                & T=1.0, TopK=2       & Y                                                                       & \textbf{0.827 }                                                                       & \textbf{0.015}         \\
\hline
\end{tabular}}
\caption{Impact of prompt duration, temperature (T), and Top-K on BnTTS-n performance in the Short-BnStudioEval Dataset.}
\label{tab:eval_on_short_Studio_eval}
\vspace{-0.3cm}
\end{table}

\noindent \textbf{Zero-shot vs. Few-shot BnTTS}: BnTTS-0 consistently falls short of BnTTS-n across all metrics in both reference-aware and reference-independent evaluations. The BnTTS-n model produces more natural and intelligible speech with high speaker fidelity, leading to improved SMOS, CER, and SECS scores. This performance gap is particularly evident in the ShortText-200 dataset, which assesses conversational fluency in short, everyday phrases. The results affirm that finetuning can significantly improve the XTTS-based model for generating natural, fluent, and speaker-adapted speech.

\noindent \textbf{High CER in Text Generation}: Both BnTTS models exhibited higher CER compared to AzureTTS and GTTS in both BnStudioEval and BnTTSTextEval datasets. The AzureTTS and GTTS also achieved a lower CER score than the GT. The BnTTS generates speech with more conversational prosody and expressiveness, which, while improving perceived quality, may negatively impact CER. ASR systems, used for CER evaluation, are often better suited to transcribing standardized speech patterns, as seen in AzureTTS and GTTS. The consistent loudness and simplified prosody in these systems create clearer phonetic boundaries, making them more easily transcribed by the ASR model \cite{Yeunju2022, Wagner2019}.

\noindent \textbf{Effect of Sampling and Prompt Length on Short Speech Generation:} 
The generation of short audio sequences presents challenges in the BnTTS models, particularly for texts containing fewer than 30 characters when using the default generation settings (Temperature \(T = 0.85\) and TopK = 50). The issues observed are twofold: (1) the generated speech often lacks intelligibility, and (2) the output speech tends to be longer than expected. To investigate this, we extracted a subset of 23 short text-speech pairs from the BnStudioEval dataset, which we call ShortBnStudioEval dataset. For evaluation, we utilize the CER metric to assess intelligibility and DurationEquality (Appendix: \ref{sec:DurationEquality}) to quantify duration discrepancies in the BnTTS-n model.

Under the default settings (Exp. 1 in Table \ref{tab:eval_on_short_Studio_eval}), the model achieves a CER of 0.081 and a DurationEquality score of 0.699. We hypothesize that this issue stems from its training process. During training, the model is accustomed to short audio prompts for short sequences. By aligning the inference with this training strategy and using short prompts, the generation performance improves vastly, as evidenced by a higher DurationEquality score of 0.820 and a lower CER of 0.029 (Exp. 2). Further, by adjusting the temperature to \(T = 1.0\) and reducing the top-K value to 2, we observed an improvement in the DurationEquality score from 0.699 to 0.701, accompanied by a substantial reduction in CER from 0.081 to 0.023 (Exp. 3). Combining the short prompt with the adjusted temperature and top-K values yielded the best results. In this configuration, the DurationEquality score improved to 0.827, with a CER of 0.015, demonstrating that both factors are crucial for accurate short speech generation.


\section{Related Works}
\label{sec:related_work}

The development of Bangla TTS technology presents unique challenges due to the language's rich morphology and phonetic diversity. The first Bangla TTS system, Katha \cite{alam2007text}, was developed using diphone concatenation within the Festival Framework. However, this approach struggled with natural prosody and efficient runtime. Later advancements, such as Subhachan \cite{naser2010implementation}, aimed to improve these aspects but still faced similar limitations. The introduction of LSTM-based models \cite{gutkin2016tts} showed promising results in Bangla speech synthesis. Beyond Bangla-specific TTS, broader efforts on Indian language synthesis have contributed to Indic-TTS systems. \citet{Prakash2020} employed Tacotron2 for text-to-mel-spectrogram conversion and WaveGlow as a vocoder. Another study \cite{indictts2022} demonstrated that monolingual models utilizing FastPitch and HiFi-GAN V1, trained on both male and female voices, outperformed previous approaches. However, these works supported a limited number of speakers and lacked speaker adaptability. To address this gap, we explore the LLM-based XTTS model for Bangla, developing the first Bangla TTS system designed for low-resource speaker adaptation.


\section{Conclusion}
In this work, we introduced BnTTS, the first speaker-adaptive TTS system for Bangla, capable of generating natural and clear speech with minimal training data. Built on the XTTS pipeline, BnTTS effectively supports zero-shot and few-shot speaker adaptation, outperforming existing Bangla TTS systems in sound quality, naturalness, and clarity. Despite its strengths, BnTTS faces challenges in handling diverse dialects and short-sequence generation. Future work will focus on training BnTTS from scratch, developing medium and small model variants, and exploring knowledge distillation to optimize inference speed for real-time applications.



\section{Limitations}
Despite the significant performance of BnTTS, the system has several limitations. It struggles to adapt to speakers with unique vocal traits, especially without prior training on their voices, limiting its effectiveness in speaker adaptation tasks. We found poor performance on short text due to pre-existing issues in the XTTS foundation model. Although we improved performance by modifying generation settings and incorporating additional training with Complete Audio Prompting, the model still fails to generate sequences under two words or 20 characters in some cases. We did not investigate the performance of the XTTS model by training from scratch; instead, we used continual pretraining due to resource constraints, which may have yielded better results.

\section{Acknowledgments}
We are grateful to HISHAB\footnote{\url{https://www.verbex.ai/}}  for providing us with
all the necessary working facilities, computational
resources, and an appropriate environment through-
out our entire work.

\section{Ethical Considerations}

The development of BnTTS raises ethical concerns, particularly regarding the potential misuse for unauthorized voice impersonation, which could impact privacy and consent. Protections, such as requiring speaker approval and embedding markers in synthetic speech, are essential. Diverse training data is also crucial to reduce bias and reflect Bangla’s dialectal variety. Additionally, synthesized voices risk diminishing dialectal diversity. As an open-source tool, BnTTS requires clear guidelines for responsible use, ensuring adherence to ethical standards and positive community impact.

\bibliography{custom}

\newpage
\appendix



\section{TTS Data Acquisition Framework}
\label{sec:data_collection}

\begin{figure}[hbt!]
    \centering
    \includegraphics[width=0.8\linewidth]{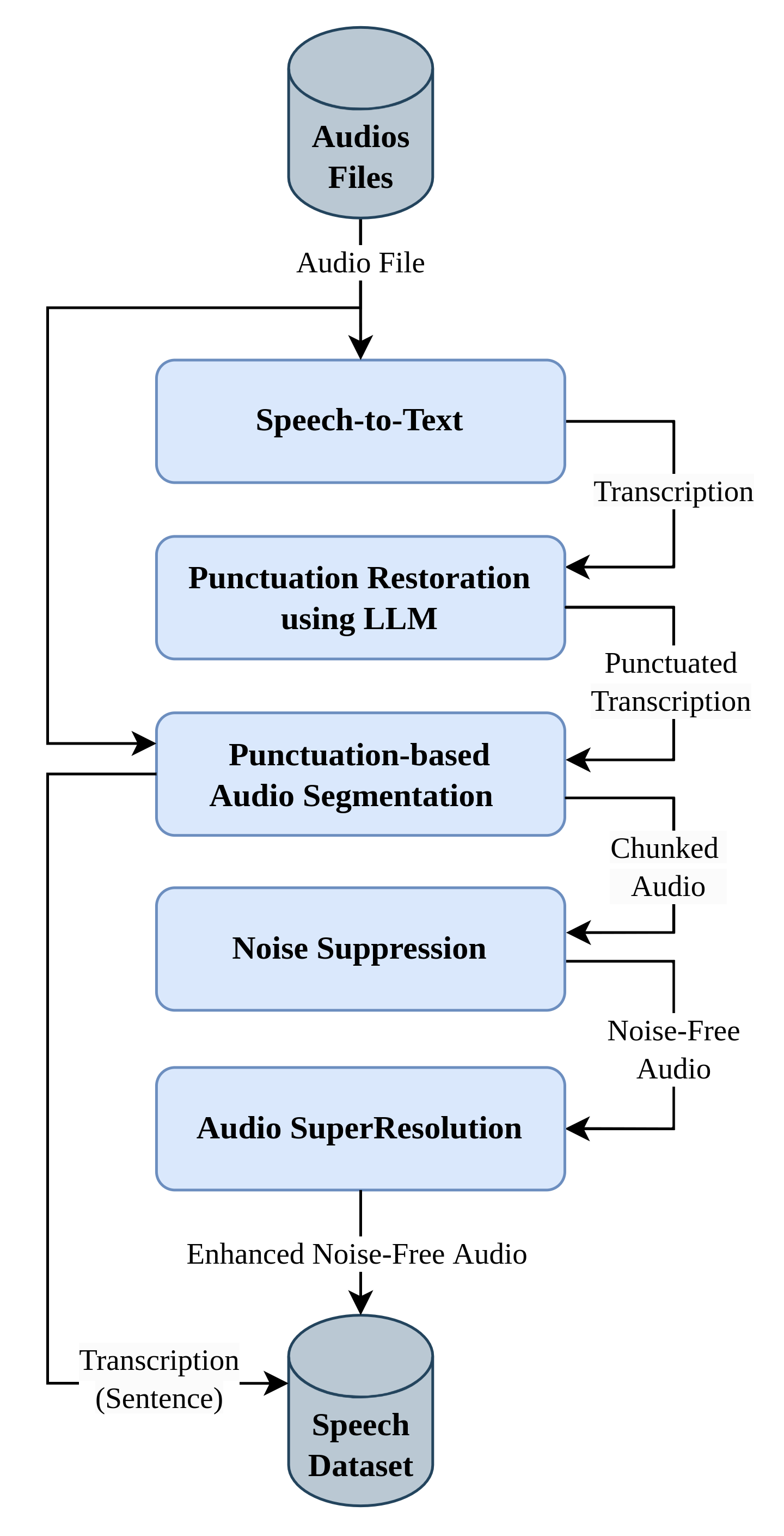} 
    \caption{Overview of our TTS Data Acquisition Framework. The acquisition process involves using a Speech-to-Text model to obtain transcription, an LLM to restore transcription's punctuation, a noise suppression model to remove unwanted noise, and finally an audio superresolution model to enhance audio quality and loudness.}
    \label{fig:pseudo_labeled_dataset}
\end{figure}

Bangla is a low-resource language, and large-scale, high-quality TTS speech data are particularly scarce. To address this gap, we developed a TTS Data Acquisition Framework (Figure \ref{fig:pseudo_labeled_dataset}) designed to collect high-quality speech data with aligned transcripts. This framework leverages advanced speech processing models and carefully designed algorithms to process raw audio inputs and generate refined audio outputs with word-aligned transcripts. Below, we provide a detailed breakdown of the key components of the framework.

\textbf{1. Speech-to-Text (STT):} The audio files are first processed through an in-house our STT system, which transcribes the spoken content into text. The STT system used here is an enhanced version of the model proposed in \cite{nandi-etal-2023-pseudo}.

\textbf{2. Punctuation Restoration Using LLM:} Following transcription, a LLM is employed to restore appropriate punctuation \cite{openai2023gpt}. This step is crucial for improving grammatical accuracy and ensuring that the text is clear and coherent, aiding in further processing.

\textbf{3. Audio and Transcription Segmentation:} The audio and transcription are segmented based on terminal punctuation (full-stop, question mark, exclamatory mark, comma). This ensures that each audio segment aligns with a complete sentence, maintaining the speaker's prosody throughout.

\textbf{4. Noise and Music Suppression:} To improve audio quality, noise and music suppression techniques \cite{defossez2019music} are applied. This step ensures that the resulting audio is free of background disturbances, which could degrade TTS performance.

\textbf{5. Audio SuperResolution:} After noise suppression, the audio files undergo super-resolution processing to enhance audio fidelity \cite{liu2021voicefixer}. This ensures high-quality audio, crucial for producing natural-sounding TTS outputs.

This pipeline effectively enhances raw audio and corresponding transcription, resulting in a high-quality pseudo-labeled dataset. By combining ASR, LLM-based punctuation restoration, noise suppression, and super-resolution, the framework can generate very high-quality speech data suitable for training speech synthesis models.

\subsection{Dataset Filtering Criteria}
\begin{figure}[hbt!]
    \centering
    \begin{subfigure}[b]{0.45\textwidth}
        \centering
        \includegraphics[width=\textwidth]{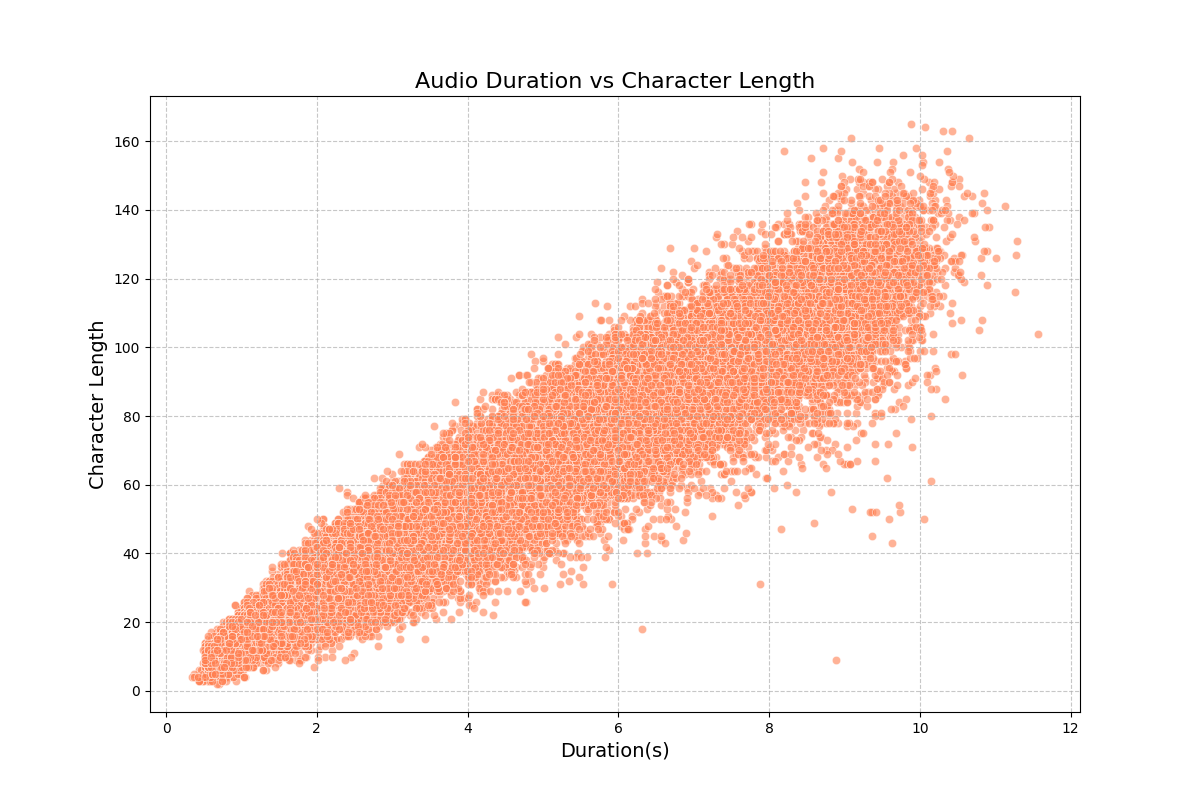}
        \caption{The diagram illustrates the linear relationship between audio duration and character length in manually-reviewed Pseudo-labeled Data - Phase A.}
        \label{fig:reviewed_data}
    \end{subfigure}
    \hfill
    \begin{subfigure}[b]{0.45\textwidth}
        \centering
        \includegraphics[width=\textwidth]{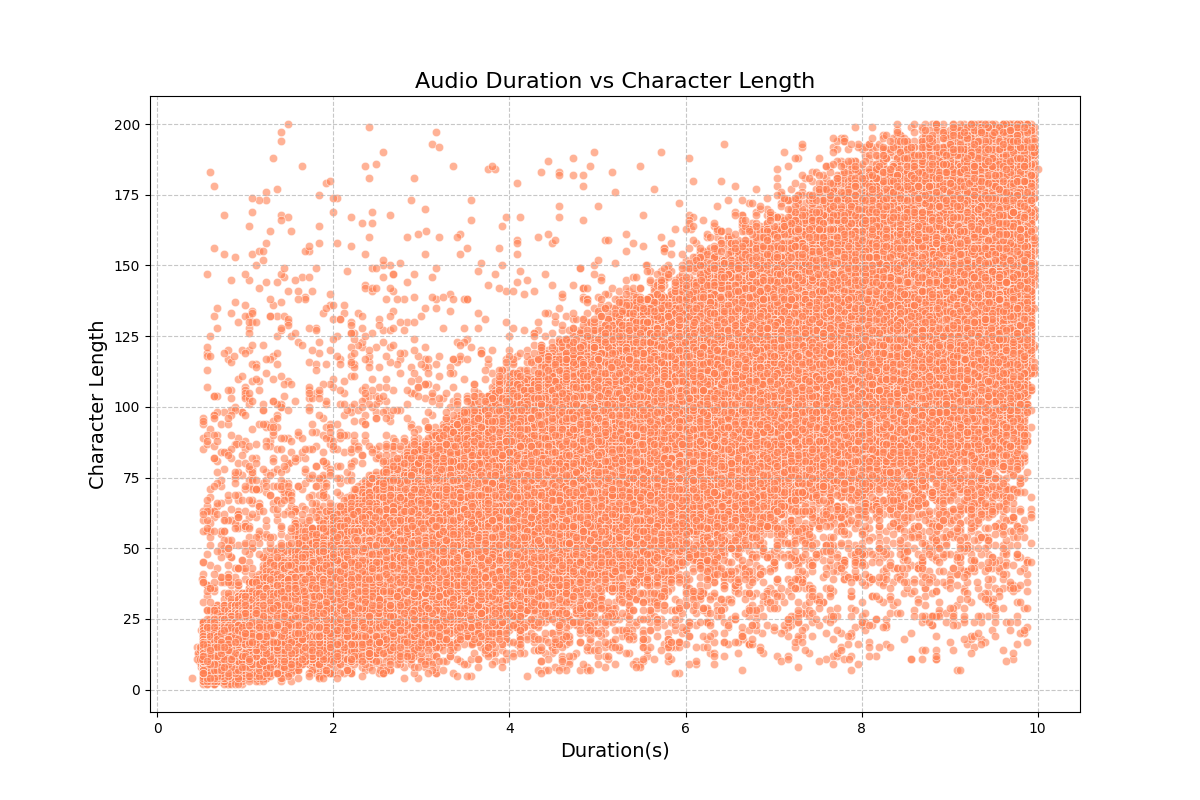}
        \caption{The diagram depicts the relationship between audio duration and character length in Pseudo-Labeled Data - Phase B.}
        \label{fig:unprocessed_data}
    \end{subfigure}
    \vskip\baselineskip
    \begin{subfigure}[b]{0.45\textwidth}
        \centering
        \includegraphics[width=\textwidth]{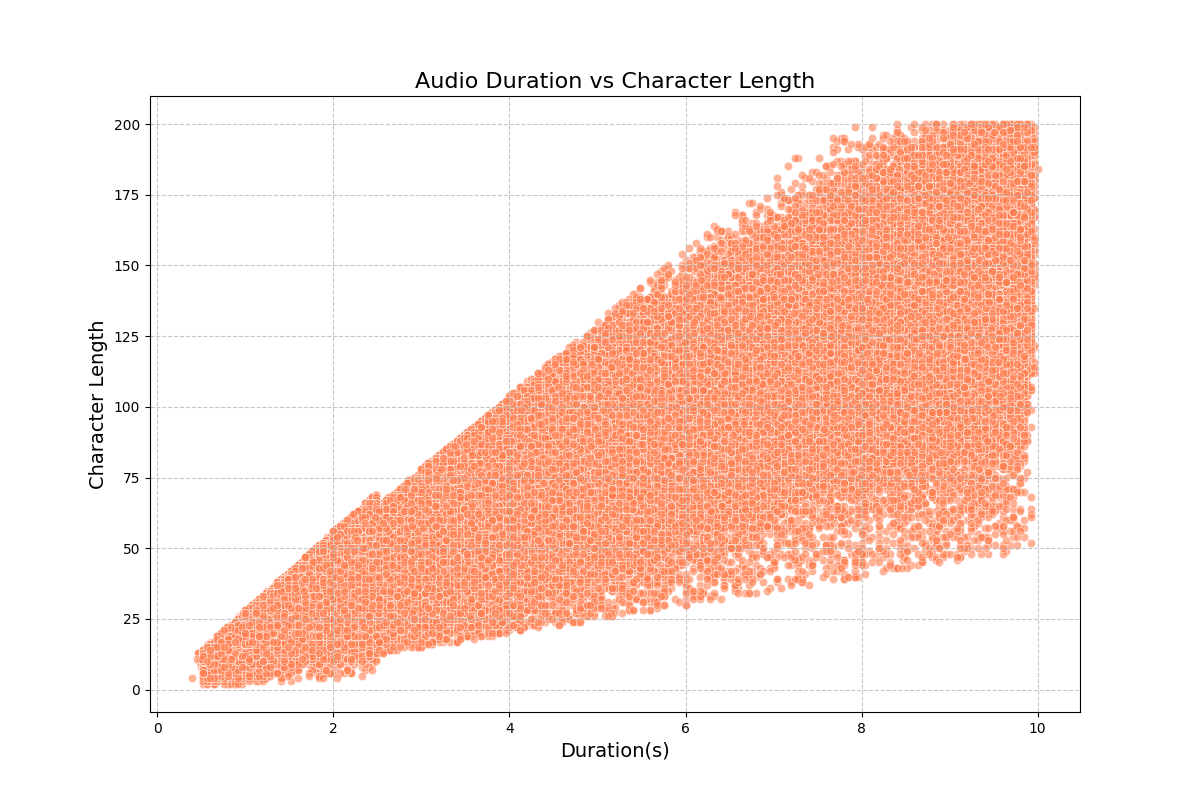}
        \caption{The diagram illustrates the audio duration vs. character length graph in Pseudo-Labeled Data - Phase B after filtering.}
        \label{fig:processed_data}
    \end{subfigure}
    \caption{These figures demonstrate how the ratio of text length to audio duration changes before and after processing the data.}
    \label{fig:audio_vs_length_grid}
\end{figure}

The pseudo-labeled data are further refined using the following criteria:

\begin{itemize}
    \item\textbf{Diarization:} Pyannote's Speaker Diarization v3.1  is employed to filter audio files by separating multi-speaker audios, ensuring that each instance contains only one speaker \cite{Plaquet23}, which is essential for effective TTS model training.

    \item \textbf{Audio Duration}: Audio segments shorter than 0.5 seconds are discarded, as they provide insufficient information for our model. Similarly, segments longer than 11 seconds are excluded to match the model’s sequence length.
    
    \item \textbf{Text Length}: Segments with transcriptions exceeding 200 characters are removed to ensure manageable input size during training.
    \item \textbf{Silence-based Filtering}: Audio files where over 35\% of the duration consists of silence are discarded, as they negatively impact model performance.
    \item \textbf{Text-to-Audio Ratio}: Based on our analysis, audio segments where the text-to-audio duration ratio falls outside (Figure \ref{fig:unprocessed_data}) the range of 6 to 25 are excluded (Figure \ref{fig:processed_data}), ensuring alignment with natural speech patterns observed in Pseudo-labeled data from Phase A (Figure \ref{fig:reviewed_data}).
\end{itemize}

\section{Human Guided Data Preparation}
\label{app:human_reviewed_data}
We curated approximately 82.39 hours of speech data through human-level observation, which we refer to as Pseudo-Labeled Data - Phase A (Table \ref{tab:dataset_info}). The audio samples, averaging 10 minutes in duration, are sourced from copyright-free audiobooks and podcasts, preferably featuring a single speaker in most cases.

Annotators were tasked with identifying prosodic sentences by segmenting the audio into meaningful chunks while simultaneously correcting ASR-generated transcriptions and restoring proper punctuation in the provided text. If a selected audio chunk contained multiple speakers, it was discarded to maintain dataset consistency. Additionally, background noise, mispronunciations, and unnatural speech patterns were carefully reviewed and eliminated to ensure the highest quality TTS training data.

\section{Dataset}
\label{app:dataset}

Table \ref{tab:dataset_info} summarizes the statistics and metadata of the datasets used in this study. We utilized four open-source datasets: OpenSLR Bangla TTS Dataset \cite{openslr_tts}, Limmits \cite{limmits24}, Comprehensive Bangla TTS Dataset \cite{com_tts}, and CRBLP TTS Dataset \cite{alam2007text}, amounting to a total of 117 hours of training data. To further enhance our dataset, we synthesized 16.44 hours of speech using Google’s TTS API, ensuring high-quality transcriptions. Additionally, 4.22 hours of professionally recorded studio speech from four speakers were collected for fine-tuning.

The majority of our dataset originates from Pseudo-Labeled Data-Phase A and Phase B. Phase A, containing 82.39 hours of speech, underwent thorough evaluation, with insights from this phase informing the refinement of the large-scale data acquisition process used in Phase B. In contrast, Phase B was generated through our TTS Data Acquisition Framework and was not manually reviewed.

\begin{table}
\centering
\footnotesize
\begin{tabular}{c|c|c} 
\hline
\textbf{Dataset}                                                           & \begin{tabular}[c]{@{}c@{}}\textbf{Duration}\\\textbf{(Hour)}\end{tabular} & \textbf{Remarks}                                                             \\ 
\hline
\begin{tabular}[c]{@{}c@{}}Pseudo-Labeled\\Data - Phase A\end{tabular}     & 82.39                                                                      & \begin{tabular}[c]{@{}c@{}}Manually\\reviewed\end{tabular}                   \\ 
\hline
\begin{tabular}[c]{@{}c@{}}Pseudo-Labeled\\Data - Phase B\end{tabular}     & 3636.47                                                                    & Not Reviewed                                                                 \\ 
\hline
Synthetic (GTTS)                                                           & 16.44                                                                      & Synthetic                                                                    \\ 
\hline
\begin{tabular}[c]{@{}c@{}}Comprehensive\\Bangla TTS\\Dataset\end{tabular} & 20.08                                                                      & \begin{tabular}[c]{@{}c@{}}Open-source\\Data\end{tabular}                    \\ 
\hline
\begin{tabular}[c]{@{}c@{}}OpenSLR Bangla\\TTS Dataset\end{tabular}        & 3.82                                                                       & \begin{tabular}[c]{@{}c@{}}Open-source\\Data\end{tabular}                    \\ 
\hline
Limmits                                                                    & 79                                                                         & \begin{tabular}[c]{@{}c@{}}Open-source\\Data\end{tabular}                    \\ 
\hline
\begin{tabular}[c]{@{}c@{}}CRBLP TTS\\Dataset\end{tabular}                 & 13.59                                                                      & \begin{tabular}[c]{@{}c@{}}Open-source\\Data\end{tabular}                    \\ 
\hline
In-House HQ Data                                                           & 4.22                                                                       & \begin{tabular}[c]{@{}c@{}}Studio Quality,\\Manually\\reviewed\end{tabular}  \\ 
\hline
\textbf{Total Duration}                                                    & \textbf{3856.01}                                                           &                                                                              \\
\hline
\end{tabular}
\caption{Dataset Information}
\label{tab:dataset_info}
\end{table}


\subsection{Evaluation Dataset}

For evaluating the performance of our TTS system, we curated two datasets: BnStudioEval and BnTTSTextEval, each serving distinct evaluation purposes.

\noindent  \textbf{BnStudioEval}: This dataset comprises 80 high-quality instances (text and audio pair) taken from our in-house studio recordings. This dataset was selected to assess the model’s capability in replicating high-fidelity speech output with speaker impersonation. 
    
\noindent \textbf{BnTTSTextEval}: The BnTTSTextEval dataset encompasses three subsets: \begin{itemize}
        \item \textbf{BengaliStimuli53}: A linguist-curated set of 53 instances, created to cover a comprehensive range of Bengali phonetic elements. This subset ensures that the model handles diverse phonemes.
        
        \item \textbf{BengaliNamedEntity1000}: A set of 1,000 instances focusing on proper nouns such as person, place, and organization names. This subset tests the model's handling of named entities, which is crucial for real-world conversational accuracy.
        \item \textbf{ShortText200}: Composed of 200 instances, this subset includes short sentences,  filler words, and common conversational phrases (less than three words) to evaluate the model’s performance in natural, day-to-day dialogue scenarios.
    \end{itemize}  

The BnStudioEval dataset, with reference audio for each text, will be for reference-aware evaluation, while BnTTSTextEval supports reference-independent evaluation. Together, these datasets provide a comprehensive basis for evaluating various aspects of our TTS performance, including phonetic diversity, named entity pronunciation, and conversational fluency.

\section{Training Objectives}
\label{app:training_objective}

Our BnTTS model is composed of two primary modules (GPT-2 and HiFi-GAN), which are trained separately. The GPT-2 module is trained using a Language Modeling objective, while the HiFi-GAN module is optimized using HiFi-GAN loss objective. This section provides an overview of the loss functions applied during training.

\subsection{Language Modeling Loss}
1. \textbf{Text Generation Loss}: Denoted as $\mathcal{L}_{\text{text}}$, it quantifies the difference between predicted logits and ground truth labels using cross-entropy. Let $\hat{y}_{\text{text}}$ represent the predicted logits and $y_{\text{text}}$ the ground truth target labels. For a sequence with $N$ text tokens, the Text Generation Loss is calculated as: 
   \begin{equation}
   \mathcal{L}_{\text{text}} = \frac{1}{N} \sum_{i=1}^{N} \text{CE}(\hat{y}_{\text{text}}^{(i)}, y_{\text{text}}^{(i)})
   \end{equation}
   
2. \textbf{Audio Generation Loss}: Denoted as $\mathcal{L}_{\text{audio}}$, it evaluates the accuracy of generated acoustic tokens against target VQ-VAE codes using cross-entropy loss:
   \begin{equation}
   \mathcal{L}_{\text{audio}} = \frac{1}{N} \sum_{i=1}^{N} \text{CE}(\hat{y}_{\text{audio}}^{(i)}, y_{\text{audio}}^{(i)})
   \end{equation}

where $\hat{y}_{\text{audio}}$ represents the predicted logits for the audio token, $y_{\text{audio}}$ are the corresponding target VQ-VAE tokens, and $N$ is the number of audio token in the sequence.
   
Total loss combines the text generation and audio generation losses with weighted factors:
   \begin{equation}
   \mathcal{L}_{\text{total}} = \alpha \mathcal{L}_{\text{text}} + \beta \mathcal{L}_{\text{audio}} \quad (\alpha = 0.01, \beta = 1.0)
   \end{equation}

where $\alpha$ and $\beta$ are scaling factors that control the relative importance of each loss term.

\subsection{HiFi-GAN Loss}
We used a HiFi-GAN-based vocoder \cite{kong2020hifi} that comprises multiple discriminators: the Multi-Period Discriminator, and Multi-Scale Discriminator. For the sake of clarity, we will refer to these discriminators as a single entity. The HiFi-GAN module is trained using multiple losses mentioned below:

1. \textbf{Adversarial Loss}: The adversarial losses for the generator \(G\) and the discriminator \(D\) are defined as follows:
\begin{align}
    \mathcal{L}_{\text{Adv}}(D; G) &= \mathbb{E}_{(x, s)} \left[(D(x) - 1)^2 + D(G(s))^2 \right] \\
    \mathcal{L}_{\text{Adv}}(G; D) &= \mathbb{E}_{s} \left[(D(G(s)) - 1)^2 \right]
\end{align}

where \(x\) represents the real audio samples, and \(s\) denotes the input conditions.

2. \textbf{Mel-Spectrogram Loss}: This loss calculates L1 distance between the mel-spectrograms of the real and generated audio. This loss is formulated as:
\begin{align}
    \mathcal{L}_{\text{Mel}}(G) = \mathbb{E}_{(x, s)} \left[\left\| \phi(x) - \phi(G(s)) \right\|_{1}\right]
\end{align}
where \(\phi\) represents the transformation function that maps a waveform to its corresponding mel-spectrogram.

3. \textbf{Feature Matching Loss}: The feature matching loss calculates the L1 distance between the intermediate features of the real and generated audio, as extracted from multiple layers of the discriminator. It is defined as:

\begin{align}
    \mathcal{L}_{\text{FM}}(G; D) = \mathbb{E}_{(x, s)} \sum_{i=1}^{T} \frac{1}{N_i} \left\| D^i(x) - D^i(G(s)) \right\|_{1}
\end{align}

where \(T\) denotes the number of discriminator layers, and \(D^i\) and \(N_i\) represent the features and number of features at the \(i\)-th layer, respectively.

\paragraph{Final Loss:}
Given that the discriminator is composed of multiple sub-discriminators, the final objectives for training the generator and the discriminator are defined as follows:
\begin{align}
    \mathcal{L}_{G} &= \sum_{k=1}^{K} \left[\mathcal{L}_{\text{Adv}}(G; D_k) + \lambda_{\text{FM}} \mathcal{L}_{\text{FM}}(G; D_k)\right] \notag \\
    &\quad + \lambda_{\text{Mel}} \mathcal{L}_{\text{Mel}}(G) \\
    \mathcal{L}_{D} &= \sum_{k=1}^{K} \mathcal{L}_{\text{Adv}}(D_k; G)
\end{align}

where \(D_k\) denotes the \(k\)-th sub-discriminator and \(\lambda_{\text{FM}} = 2\), \(\lambda_{\text{Mel}} = 45\).





\section{Evaluation Metrics}
\label{app:eval_metrics}
We employed a combination of subjective and objective metrics to rigorously evaluate the performance of our TTS system, focusing on intelligibility, naturalness, speaker similarity, and transcription accuracy.

\noindent \textbf{Subjective Mean Opinion Score (SMOS):} SMOS is a perceptual evaluation where listeners rate synthesized speech on a Likert scale from 1 (poor) to 5 (excellent). It considers naturalness, clarity, and fluency, providing an absolute score for each sample. A higher SMOS indicates better overall speech quality.

\noindent \textbf{SpeechBERTScore:} SpeechBERTScore adapts BERTScore for speech, using self-supervised learning (SSL) models to compare dense representations of generated and reference speech. For generated speech waveform $\hat{X}$ and reference waveform $X$, the feature representations $\hat{Z}$ and $Z$ are extracted using a pretrained model. SpeechBERTScore is defined as the average maximum cosine similarity between feature vectors:
\[
\text{SpeechBERTScore} = \frac{1}{N_{\text{gen}}} \sum_{i=1}^{N_{\text{gen}}} \max_{j} \text{cos}(\hat{\mathbf{z}}_i, \mathbf{z}_j)
\]
where $\hat{\mathbf{z}}_i$ and $\mathbf{z}_j$ represent the SSL embeddings for generated and reference speech, respectively.

\noindent \textbf{Character Error Rate (CER):} CER measures transcription accuracy by calculating the ratio of errors (substitutions $S$, deletions $D$, and insertions $I$) in automatic speech recognition (ASR) transcriptions:
\[
CER = \frac{S + D + I}{N}
\]
where $N$ is the total number of characters in the reference transcription. A lower CER indicates better transcription accuracy.

\noindent \textbf{Speaker Encoder Cosine Similarity (SECS):} SECS evaluates speaker similarity by calculating the cosine similarity between speaker embeddings of the reference and synthesized speech:

\[
\text{SECS} = \frac{e_{\text{ref}} \cdot e_{\text{syn}}}{\|e_{\text{ref}}\| \|e_{\text{syn}}\|},
\]

where $e_{\text{ref}}$ and $e_{\text{syn}}$ are the speaker embeddings for reference and synthesized speech, respectively. SECS ranges from -1 (low similarity) to 1 (high similarity).

\label{sec:DurationEquality}
\noindent \textbf{Duration Equality Score:} This metric quantifies how closely the durations of the reference ($a$) and synthesized ($b$) speech match, with a score of 1 indicating identical durations:

\[
\text{DurationEquality}(a, b) = \frac{1}{\max\left(\frac{a}{b}, \frac{b}{a}\right)}.
\]

This score helps in assessing duration similarity between reference and generated audio, ensuring consistency in pacing.

Each metric provides a different perspective, allowing a holistic evaluation of the synthesized speech quality.

\section{Subjective Evaluation}
For subjective evaluation of our system, we employ the Mean Opinion Score (MOS), a widely recognized metric primarily focusing on assessing the perceptual quality of audio outputs. To ensure the reliability and accuracy of our evaluations, we carefully select a panel of ten experts who are thoroughly trained in the intricacies of MOS scoring. These experts are equipped with the necessary skills and knowledge to critically assess and score the system, providing invaluable insights that help guide the refinement and enhancement of our technology. This structured approach guarantees that our evaluations are both comprehensive and precise, reflecting the true quality of the audio outputs under review.

\subsection{Evaluation Guideline}
For calculating MOS, we consider five essential evaluation criteria:
\begin{itemize} \item \textbf{Naturalness:} Evaluates how closely the TTS output resembles natural human speech. \item \textbf{Clarity:} Assesses the intelligibility and clear articulation of the spoken words. \item \textbf{Fluency:} Examines the smoothness of speech, including appropriate pacing, pausing, and intonation. \item \textbf{Consistency:} Checks the uniformity of voice quality across different texts. \item \textbf{Emotional Expressiveness:} Measures the ability of the TTS system to convey the intended emotion or tone. \end{itemize}

In the evaluation, we employ a five-point rating scale to meticulously assess performance based on specific criteria. This scale ranges from 1, denoting 'Bad' where the output has significant distortions, to 5, representing 'Excellent' where the output nearly replicates natural human speech and excels in all evaluation aspects. To capture more subtle nuances in the TTS output that might not perfectly fit into these whole-number categories, we also recommend using fractional scores. For example, a 1.5 indicates quality between 'Bad' and 'Poor,' a 2.5 signifies improvement over 'Poor' but not quite reaching 'Fair,' a 3.5 suggests better than 'Fair' but not up to 'Good,' and a 4.5 reflects performance that surpasses 'Good' but falls short of 'Excellent.' This fractional scoring allows for a more precise and detailed reflection of the system's quality, enhancing the accuracy and depth of the MOS evaluation.

\subsection{Evaluation Process}
We have developed an evaluation platform specifically designed for the subjective assessment of Text-to-Speech (TTS) systems. This platform features several key attributes that enhance the effectiveness and reliability of the evaluation process. Key features include anonymity of audio sources, ensuring that evaluators are unaware of whether the audio is synthetically generated or recorded from studio environment, or which TTS model, if any, was used. This promotes unbiased assessments based purely on audio quality. Comprehensive evaluation criteria allow evaluators to rate each audio sample on naturalness, clarity, fluency, consistency, and emotional expressiveness, ensuring a holistic review of speech synthesis quality. The user-centric interface is streamlined for ease of use, enabling efficient playback of audio samples and score entry, which reduces evaluator fatigue and maintains focus on the task. Finally, the structured data collection method systematically captures all ratings, facilitating precise analysis and enabling targeted improvements to TTS technologies. This platform is a vital tool for developers and researchers aiming to refine the effectiveness and naturalness of speech outputs in TTS systems.

\subsection{Evaluator Statistics}
For our evaluation process, we carefully selected 10 expert native speakers, achieving a balanced representation with 5 males and 5 females. The age range for these evaluators is between 20 to 28 years, ensuring a youthful perspective that aligns well with our target demographic. All evaluators are either currently enrolled as graduate students or have already completed their graduate studies. They hail from a variety of academic backgrounds, including economics, engineering, computer science, and social sciences, which provides a diverse range of insights and expertise. This careful selection of qualified individuals ensures a comprehensive and informed assessment process, suitable for our needs in evaluating advanced systems or processes where diverse, educated opinions are crucial.

\subsection{Subjective Evaluation Data Preparation} 
For reference-aware evaluation, we selected 20 audio samples from each of the four speakers, resulting in 80 Ground Truth (GT) audios. To facilitate comparison, we generated 400 synthetic samples (80 × 5) using the TTS systems examined in this study. Including the GT samples, the total dataset for this evaluation amounts to 480 audio files (400 + 80).

For the reference-independent evaluation, we utilized 453 text samples from BnTTSTextEval, comprising BengaliStimuli53 (53), BengaliNamedEntity1000 (200), and ShortText200 (200). Given the four speakers in both BnTTS-0 and BnTTS-n, this resulted in 3,624 audio samples (4 × 453 × 2). Additionally, IndicTTS, GTTS, and AzureTTS contributed 1,359 samples (3 × 453). IndicTTS samples were evenly distributed between two male and female speakers, while GTTS and AzureTTS used the "bn-IN-Wavenet-C" and "bn-IN-TanishaaNeural" voices, respectively.

In total, the reference-independent evaluation dataset comprised 5,436 audio samples. When combined with the 480 samples from the reference-aware evaluation, the overall dataset for subjective evaluation amounted to 5,916 audio files. These samples were randomly mixed and distributed to the reviewer team to ensure unbiased evaluations.

\section{Use of AI assistant}
\label{sec:use_of_ai_assistant}
We used AI assistants such as GPT-4o for spelling and grammar checking for the text of the paper.

\newpage


\section{Symbols and Notations}
\label{sec:notation}

\begin{table}[H]
\centering
\setlength{\tabcolsep}{2.9pt}
\resizebox{0.45\textwidth}{!}{%

    \begin{tabular}{c|l}
        \hline
        \textbf{Variable} & \textbf{Description} \\ \hline
        \( \mathbf{T} \) & Text sequence with \( N \) tokens \\ \hline
        \( N \) & Number of tokens in the text sequence \\ \hline
        \( \mathbf{S} \) & Speaker's mel-spectrogram with \( L \) frames \\ \hline
        \( \hat{\mathbf{Y}} \) & Generated speech  \\ \hline
        \( \mathbf{Y} \) & Ground truth mel-spectrogram  \\ \hline
        \( \mathcal{F} \) & LLM Model \\ \hline
        \( \mathbf{z} \) & Discrete codes \\ \hline
        \( \mathcal{C} \) & Codebook of discrete codes \\ \hline
        \( l \) & Number of layers \\ \hline
        \( k \) & Number of attention heads i \\ \hline
        \( \mathbf{S_z} \) & speaker spectrogram embd.\( \mathbb{R}^{L \times d} \) \\ \hline
        \( d \) & Embedding \\ \hline
        \( \mathbf{Q}, \mathbf{K}, \mathbf{V} \) & Query, Key, Value \\ \hline
        \( P \) & Perceiver Resampler \\ \hline
        \( \mathbf{R} \) & Fixed-size output in \( \mathbb{R}^{P \times d} \) \\ \hline
        \( \mathbf{T_e} \) & Continuous embedding  \( \mathbb{R}^{N \times d} \) \\ \hline
        \( \mathbf{S_p} \) & Speaker embeddings \\ \hline
        \( \mathbf{Y_z} \) & Ground truth  \\ \hline
        \( \mathbf{X} \) & Input of LLM   \\ \hline
        \( \oplus \) & Concatenation operation \\ \hline
        \( \mathbf{H} \) & Output from the LLM  \\ \hline
        \( \mathbf{H}_\text{Y} \) & Spectrogram embedding  \\ \hline
        \( \mathbf{S}' \) & Resized embedding \( \mathbf{H}_\text{Y} \) \\ \hline
        \( \mathbf{W} \) & Final audio waveform \\ \hline
        \( g_\text{HiFi} \) & HiFi-GAN function \\ \hline
    \end{tabular}}
    \label{tab:variables_descriptions}
    \caption{Table of Variables and Descriptions}
\end{table}

\end{document}